\documentclass[11pt,a4paper]{article}
\usepackage[hyperref]{acl2021}
\usepackage{times}
\usepackage{latexsym}

\usepackage[ruled,vlined]{algorithm2e}
\usepackage{graphicx} % DO NOT CHANGE THIS
\usepackage{bm}
\usepackage{booktabs}
\usepackage{microtype}

\aclfinalcopy % Uncomment this line for the final submission
\def\aclpaperid{3088} %  Enter the acl Paper ID here

\newcommand{\sysname}{NSQA}
\newcommand{\sparql}{\textsc{sparql}\xspace}
\newcommand{\ask}{\textsc{ask}\xspace}
\newcommand{\where}{\textsc{where}\xspace}
\newcommand{\select}{\textsc{select}\xspace}
\newcommand{\countsparql}{\textsc{count}\xspace}
\newcommand{\sortsparql}{\textsc{sort}\xspace}

\usepackage{tikz}
\def\checkmark{\tikz\fill[scale=0.4](0,.35) -- (.25,0) -- (1,.7) -- (.25,.15) -- cycle;} 

\title{Leveraging Abstract Meaning Representation for Knowledge Base Question Answering}

\author{
    Pavan Kapanipathi\thanks{~Equal contribution, correspondence to Pavan Kapanipathi (kapanipa@us.ibm.com), Ibrahim Abdelaziz (ibrahim.abdelaziz1@ibm.com), Srinivas Ravishankar 
(srini@ibm.com)},~Ibrahim Abdelaziz\footnotemark[1],~Srinivas Ravishankar\footnotemark[1], Salim Roukos, \\ 
    \textbf{Alexander Gray, Ramon Astudillo, Maria Chang, Cristina Cornelio, Saswati } \\ \textbf{Dana, Achille Fokoue, Dinesh Garg, Alfio Gliozzo, Sairam Gurajada, Hima}  \\
    \textbf{Karanam, Naweed Khan, Dinesh Khandelwal, Young-Suk Lee, Yunyao Li, }  \\
    \textbf{Francois Luus, Ndivhuwo Makondo, Nandana Mihindukulasooriya, Tahira} \\ \textbf{ Naseem, Sumit Neelam, Lucian Popa, Revanth  Reddy, Ryan Riegel, Gaetano} \\ \textbf{ Rossiello, Udit Sharma, G P Shrivatsa Bhargav, Mo Yu} \\
    IBM Research
}

\begin{document}
% \linenumbers
\maketitle
%--------Main paper---------------
\begin{abstract}
%Knowledge base question answering (KBQA) is a task where end-to-end deep learning techniques have faced significant challenges such as the need for semantic parsing, reasoning, domain adaptation, and large training datasets. 
Knowledge base question answering (KBQA) is an important task in Natural Language Processing. Existing approaches face significant challenges including complex question understanding, necessity for reasoning, and lack of large end-to-end training datasets. 
%On the other hand, most approaches that solve KBQA on small datasets are graph driven with minimal focus on natural language question understanding. 
%, making rapid domain adaptation extremely difficult.
In this work, we propose Neuro-Symbolic Question Answering (\sysname), a modular KBQA system, that leverages (1) Abstract Meaning Representation (AMR) parses for task-independent question understanding; (2) a simple yet effective graph transformation approach to convert AMR parses into candidate logical queries that are aligned to the KB; 
% (3) a neuro-symbolic reasoner called Logical Neural Network (LNN) that executes logical queries and reasons over KB facts to provide an answer; 
(3) a pipeline-based approach  which integrates multiple, reusable modules that are trained specifically for their individual tasks (semantic parser, entity and relationship linkers,  and neuro-symbolic reasoner) and do not require end-to-end training data.
% \sysname~utilizes a novel path-based approach to transform AMR parses of  questions into candidate logical queries aligned to the knowledge base and are used by LNN to answer the question. 
%, hence requires  minimal re-training for the overall KBQA task 
 {\sysname}~ achieves state-of-the-art performance on two prominent KBQA datasets based on DBpedia (QALD-9 and LC-QuAD 1.0). Furthermore, our analysis emphasizes that AMR is a powerful tool for KBQA systems.
 
 %\sysname's novelty lies in its modular neuro-symbolic architecture and its task-general approach to interpreting natural language questions. %The novelty lies in the fact that {\sysname} system is \textcolor{red}{general} and produces state-of-the-art performance on \texttt{QALD-9} and \texttt{Lc-QuAD-1} with minimal fine tuning.  showing 

\end{abstract}
\section{Introduction}

\begin{figure*}
  \centering
    \includegraphics[width=1\textwidth]{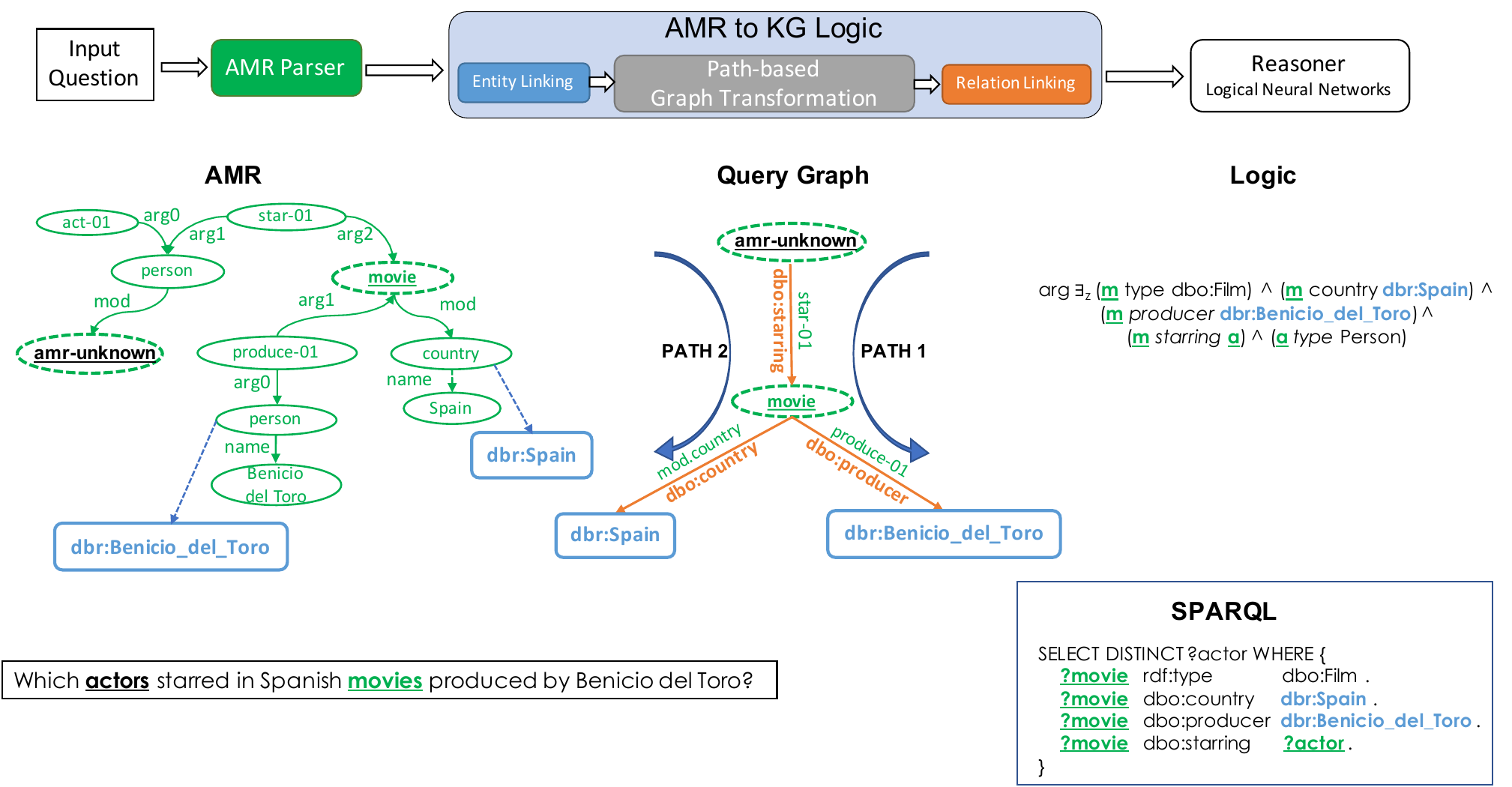}
    \caption{Real {\sysname} prediction for the sentence \textit{Which actors starred in Spanish movies produced by Benicio del Toro?}. In underlined, we show the representation for the two unknown variables across all stages including: AMR-aligned tokens in sentence (\textit{Which}, \textit{movies}), AMR graph (\textit{unknown}, \textit{movie}), paths representation (same as AMR), logical representation (\textit{actor}, \textit{movie}) and SPARQL interpretation (\textit{?actor}, \textit{?movie}). Displayed stage outputs: AMR (green), Entity Linking (blue), Relation Linking (orange)} 
  \label{fig:arch}
\end{figure*}

Knowledge base question answering (KBQA) is a sub-field within Question Answering with  desirable characteristics for real-world applications. KBQA requires a system to answer a natural language question based on facts available in a Knowledge Base (KB)~\cite{zou2014natural,vakulenko2019message,diefenbach2020towards,abdelaziz2021semantic}. Facts are retrieved from a KB through structured queries (in a query language such as \sparql), which often contain multiple triples that represent the steps or antecedents required for obtaining the answer. 
%This enables a transparent and self-explanatory form of QA, which is open domain within the limitations of the KB. 
This enables a transparent and self-explanatory form of QA, meaning that intermediate symbolic representations capture some of the steps from natural language question to answer. 

%The scope and structure of the KB is an important feature for many applications, enabling humans to control the knowledge explicitly, e.g. to ensure compliance with regulatory frameworks or to add new knowledge incrementally.

% \textcolor{red}{KBQA enables easy human intervention and oversight e.g. to ensure compliance with regulatory frameworks, but also to enrich or expand available knowledge in an incremental manner.}

With the rise of neural networks in NLP,  various KBQA models approach the task in an end-to-end manner.
%However, end-to-end models lose desirable properties, such as transparently and faithfully representing the underlying operation logic \cite{sun2020faithful}. Other systems require external fact memories to retain traceability of decisions \cite{verga2020facts}. %{\color{red}An end-to-end model looses however desirable properties, such being faithful to the underlying operation logic \cite{sun2020faithful}. Or requires to add external fact memories to retain traceability of decisions \cite{verga2020facts}.} 
Many of these approaches formulate text-to-query-language as sequence-to-sequence problem, and thus require sufficient examples of paired natural language and target representation pairs. 
% \begin{figure*}[t]
%   \centering
%     \includegraphics[width=1\textwidth]{images/nsqa_diagram2.pdf}
%     \caption{Real {\sysname} prediction for the sentence \textit{Which actors starred in Spanish movies produced by Benicio del Toro?}. In underlined, we show the representation for the two unknown variables across all stages including: AMR-aligned tokens in sentence (\textit{actors}, \textit{movies}), AMR graph (\textit{unknown person}, \textit{movie}), paths representation (same as AMR), logical representation (\textit{a}, \textit{m}) and SPARQL interpretation (\textit{?actor}, \textit{?movie}). Displayed stage outputs: AMR (green), Entity Linking (blue), Relation Linking (orange)} 
%   \label{fig:arch}
% \end{figure*}
%~\cite{keysers2019measuring,lake2018generalization}
% sun2020sparqa https://arxiv.org/pdf/2003.13956.pdf has nice related work on SP for KBQA
However, labeling large amounts of data for KBQA is challenging, either due to the requirement of expert knowledge~\cite{qald7}, or artifacts introduced during automated creation~\cite{lcquad}. 
Real-world scenarios require solving complex multi-hop questions i.e. secondary unknowns within a main question and questions employing unusual expressions. Pipeline approaches can delegate language understanding to pre-trained semantic parsers, which mitigates the data problem, but are considered to suffer from error propagation. %Solutions like the use of simplified, more robust, representations \cite{sun2020sparqa,wolfson2020break} remove some of the rich information of sematic parses, making the process less transparent. 
%Some solutions reduce the difficulty of the semantic parsing task by making use of simplified representations that are more robust to the complexities of natural language questions \cite{sun2020sparqa,wolfson2020break}, but lack the expressivity, generality, and transparency of rich semantic parses. 
However, the performance of semantic parsers for well-established semantic representations has greatly improved in recent years. Abstract Meaning Representation (AMR)~\cite{banarescu2013abstract, dorr1998thematic} parsers recently reached above $84$\% F-measure~\cite{bevilacqua2021one}, an improvement of over $10$ points in the last three years. 
%Furthermore, general domain and domain specific data is available to train English AMR parsers as well as other languages such as Chinese and Spanish.

In this paper we propose Neuro-Symbolic Question Answering (NSQA), a modular knowledge base question answering system with the following objectives: (a) delegating the complexity of understanding natural language questions to AMR parsers; (b) reducing the need for end-to-end (text-to-\sparql) training data with a pipeline architecture where each module is trained for its specific sub-task; (c) facilitating the use of an independent reasoner via an intermediate logic form. 
%retaining the transparency and traceability of traditional KBQA models while enabling understanding of complex natural queries;  

The contributions of this work are as follows:
\begin{itemize}
    \item The first system to use Abstract Meaning Representation for KBQA achieving state of the art performance on two prominent datasets on DBpedia (QALD-9 and LC-QuAD 1.0).
    \item A novel, simple yet effective path-based approach that transforms AMR parses into intermediate logical queries that are aligned to the KB. This intermediate logic form facilitates the use of neuro-symbolic reasoners such as Logical Neural Networks~\cite{riegel2020logical}, paving the way for complex reasoning over knowledge bases. 
    \item A pipeline-based modular approach that integrates  multiple, reusable modules that are trained specifically for their individual tasks (e.g. semantic parsing,  entity  linking, and  relationship  linking)  and  hence do  not  require  end-to-end training data. 
    % \item 4. Paving the way for seamlessly incorporating increasingly capable reasoners via an intermediate logic form. {\color{red} Write about transforming to logic and use of reasoner as a module.}
\end{itemize}

\section{Approach Overview}
\label{sec:approach}

% In this section, we first provide an overview of {\sysname} and detail each of its components in the following sections.

%$\arg\exists_z$(($x$ \texttt{type Film}) $\land$ ($x$ \texttt{country Spain}) \\ ($x$ \texttt{producer Benicio\_del\_Toro}) $\land$ ($x$ \texttt{starring} $z$) $\land$ ($z$ \texttt{type Person}))
% \\

% \begin{itemize}
% \item Semantic Parser
% \item Extended AMR Generator
% \begin{itemize}
% \item Entity Linking
% \item Path based triple generation 
% \item Relation Linking 
% \end{itemize}
% \item Semantic Parse to Logic Translator  
% \item Logical Neural Network as Reasoner
% \end{itemize}

Figure~\ref{fig:arch} depicts the pipeline of our {\sysname} system. Given a question in natural language, \sysname:  (i) parses questions into an Abstract Meaning Representation (AMR) graph; % exploiting advances in general language understanding independent of the KBQA task;
%ii) Links elements in the AMR graph to entities and relations in DBpedia, iii) translates the knowledge-based enhanced AMR graph into a logical query, and 
 (ii) transforms the AMR graph to a set of candidate KB-aligned logical queries, via a novel but simple graph transformation approach; 
%(iii) these logical queries enable the use of existing logical reasoners, such as Logical Neural Network (LNN)~\cite{riegel2020logical}, capable of performing explicit symbolic reasoning over KB facts to produce answers. %\sysname's modules operate within a pipeline and are individually trained for their specific tasks. 
(iii) uses a Logical Neural Network (LNN)~\cite{riegel2020logical} to reason over KB facts and produce answers to KB-aligned logical queries.
We describe each of these modules in the following sections.

%, which begins with a question in natural language and ends with the answer(s) retrieved from the KG. We take the traditional approach for knowledge base question answering by semantic parsing 

\subsection{AMR Parsing}

%In general, concepts can be implicitly related to one or more surface symbols but there are no explicit alignments between concepts and surface symbols. 
%Current state of the art systems for AMR leverage either transition-based \cite{naseem2019rewarding} or graph-based \cite{zhang2019broad} parsing approaches, parameterized with neural networks and trained in end-to-end fashion. 
%The human-annotated question data set, separate from the AMR3.0 train set, contains 877 sentences (details in Section \ref{sec:evaluation}). %includes 250 questions from the QALD dataset (a subset of the entire QALD dataset) and 11k questions from LC-QUAD. 
%The pre-trained AMR model is fine-tuned on the human-annotated questions only, significantly improving the AMR parser performance on QALD and LC-QUAD domain.
% \textcolor{red}{Example of AMR and explain Propbank and amr-unknown here.}
%As depicted in the two examples in the left hand side of Fig.~\ref{fig:arch}, 

%In order to deal with queries of arbitrary complexity and input noise, {\sysname} utilizes AMR parsing. 
{\sysname} utilizes AMR parsing to reduce the complexity and noise of natural language questions.
An AMR parse is a rooted, directed, acyclic graph. AMR nodes represent concepts, which may include normalized surface symbols, Propbank frames \cite{kingsbury2002treebank} as well as other AMR-specific constructs to handle named entities, %linked entities, 
quantities, dates and other phenomena. Edges in an AMR graph represent the relations between concepts such as standard OntoNotes roles but also AMR specific relations such as polarity or mode.

As shown in Figure \ref{fig:arch}, AMR provides a representation that is fairly close to the KB representation. A special \textit{amr-unknown} node, indicates the missing concept that represents the answer to the given question. In the example of Figure \ref{fig:arch}, \textit{amr-unknown} is a  \textit{person}, who is the subject of \textit{act-01}. %(acting in the \textit{film/theater actor} sense). 
Furthermore, AMR helps identify intermediate variables that behave as secondary unknowns.
% for questions that have more than one unknown, like the one displayed, AMR also factors out the context necessary to identify the secondary unknown.
In this case, a \textit{movie} produced by \textit{Benicio del Toro} in \textit{Spain}.

\sysname~ utilizes a stack-Transformer transition-based model \cite{naseem2019rewarding,fernandez2020a} for AMR parsing. 
% This employs a modified sequence to sequence Transformer with hard attention on top of RoBERTa \cite{liu2019roberta}. 
An advantage of transition-based systems is that they provide explicit question text to AMR node alignments. 
This allows encoding closely integrated text and AMR input to multiple modules (Entity Linking and Relation Linking) that can benefit from this joint input.
% This allows to track final system decisions back into the original tokens in the input sentence.

%Entity linking in AMR is delegated to the system detailed in the next section.

%We use an off-the-shelve implementation of the stack-Transformer \footnotemark\footnotetext{\url{https://github.com/IBM/transition-amr-parser}} trained with the official AMR3.0 dataset and self-learning techniques following \cite{youngsuklee-etal-emnlp2020} and a small adaptation set for the target domain (details in Section \ref{sec:evaluation}).

%\subsection{Abstract Meaning Representation to Logic}
\subsection{AMR to KG Logic}
The core contribution of this work is our next step where the AMR of the question is transformed to a query graph aligned with the underlying knowledge graph. We formalize the two graphs as follows:

\vspace{0.5em}
\noindent \textbf{AMR graph $\mathcal{G}$} is a rooted  edge-labeled directed acyclic graph $\langle \mathcal{V_G, E_G} \rangle $. The edge set $\mathcal{E_G}$ consists of non-core roles, quantifiers, and modifiers. The vertex set $\mathcal{V_G} \in$ \textit{amr-unknown} $\cup~\mathcal{A_P}$ $\cup$ $\mathcal{A_C}$ where where $\mathcal{A_P}$ are set of propbank predicates and $\mathcal{A_C}$ are rest of the  nodes\footnote{\url{https://www.isi.edu/~ulf/amr/help/amr-guidelines.pdf}}. Propbank predicates are \textit{n-ary} with multiple edges based on their definitions. \textit{amr-unknown} is a special concept node in the AMR graph indicating \textit{wh-questions}.
%We separate the definition of $amr-unknown$ plays a critical role for our question answering system. In cases of imperative and interrogative questions that does not contain an $amr-unknown$, {\color{red}we introduce it based on the heuristics mentioned in Algorithm 1.}
% $\mathcal{E_G}$ denote relationships between these nodes. 

Further, we enrich the AMR Graph $\mathcal{G}$ with explicit links to entities in the KG. 
For example,  the question in Figure 1 contains two entities \textit{Spain} and \textit{Benicio Del Toro} that need to be identified and linked to DBpedia entries,  \textit{dbr:Spain} and \textit{dbr:Benicio\_del\_toro}. 
%{\color{red}{This is a mapping from $\mathcal{A_C_E} \rightarrow \mathcal{V_E}$ where $\mathcal{V_E}$ is the set of entities in the underlying KG and $\mathcal{A_C_E} \subset \mathcal{A_C}$}}. 
Linking these entities is absolutely necessary for any KBQA system~\cite{zou2014natural,vakulenko2019message}. To do so, we trained a BERT-based neural mention detection model and used  BLINK~\cite{DBLP:journals/corr/abs-1810-04805} for disambiguation. The entities are linked to AMR nodes based on the AMR node-text alignment information.
The linking is a bijective mapping from $\mathcal{V}_e \rightarrow E$ where $\mathcal{V}_e$ is the set of AMR entity nodes, and $E$ is the set of entities in the underlying KG.

\vspace{0.5em}
\noindent \textbf{Query graph $\mathcal{Q}$} is a directed edge-labeled graph $\mathcal{\langle V_Q, E_Q \rangle}$, which has a similar structure to the underlying KG. $\mathcal{V_Q} \in \mathcal{V_E} \cup \mathcal{V}$ where $\mathcal{V_E}$ is a set of entities in the KG and ($\mathcal{V}$) is a set of unbound variables. $\mathcal{E_Q}$ is a set of binary relations among $\mathcal{V_Q}$ from the KG. The Query Graph $\mathcal{Q}$ is essentially the \where clause\footnote{The Query Graph does not include the type constraints in the \sparql \where Clause.} in the \sparql query. 
%The direction of the edges will be resolved during relation linking in the next module, to match the underlying KG structure.

%Our next step is to transform the AMR graph of the question to a query graph that aligns to the underlying KG and can be used in the WHERE clause of the final SPARQL query. %This forms the core of our approach and the basis of leveraging AMR-graphs for KBQA. 

% \input{sections/EL}
% \input{sections/REL}
% \input{sections/AMR2Logic}

Our goal is to transform the AMR graph $\mathcal{G}$ into its corresponding query graph $\mathcal{Q}$. However such transformation faces the following challenges:

\noindent - \textbf{N-ary argument mismatch}: Query graph $\mathcal{Q}$ represents information using binary relations, whereas AMR graph contain Propbank framesets that are n-ary. For example, the node  \textit{produce-01}\footnote{\url{http://verbs.colorado.edu/propbank/framesets-english-aliases/produce.html}} from $\mathcal{A_P}$ in $\mathcal{G}$ has four possible arguments, whereas its corresponding KG relation in $\mathcal{Q}$ (\textit{dbo:producer}) is a binary relation. 

\noindent - \textbf{Structural and Granular mismatch}: %AMR is represented as a graph where predicates may be represented as edges or nodes. 
    The vertex set of the query graph $\mathcal{Q}$ represent entities or unbound variables. On the other hand, AMR Graph $\mathcal{G}$ contains nodes that are concepts or PropBank predicates which can correspond to both entities and relationships. For example in Figure~\ref{fig:arch}, \textit{produce-01, star-01,} and \textit{Spain} are nodes in the AMR graph. So the AMR graph  $\mathcal{G}$ has to be transformed such that nodes primarily correspond to entities and edges (edge labels) correspond to relationships. Furthermore, it is possible for multiple predicates and concepts from $\mathcal{G}$ to jointly represent a single binary relation in $\mathcal{Q}$ because the underlying KG uses a completely different vocabulary. An example of such granular mismatch is shown in  Figure~\ref{fig:cocoa_bean}.

\subsubsection{Path-based Graph Transformation}

We address the challenges mentioned above by using a path-based approach for the construction of Query Graphs. In KBQA, query graphs (i.e. \sparql queries) constrain the unknown variable based on paths to the grounded entities. In Figure~\ref{fig:arch}, the constraints in the \sparql query are based on paths from \textit{?actor} to \textit{dbr:Benicio\_del\_toro} and \textit{dbr:Spain} as shown below.

{\small
\begin{itemize}\setlength\itemsep{-0.1em}
\item \texttt{?actor $\rightarrow$ dbo:starring $\rightarrow$ ?movie $\rightarrow$ \\ dbo:country $\rightarrow$ dbr:Spain}
\item \texttt{?actor $\rightarrow$ dbo:starring $\rightarrow$ ?movie $\rightarrow$ dbo:producer $\rightarrow$ dbr:Benicio del Toro}
\end{itemize}
}

% Therefore, in our first step we introduce the grounded entities to the AMR graph as described below.

%\input{sections/EL_new}

%Following the intuition of path-based constraints in SPARQL queries, we address the above challenges by a path-based approach that 

Based on this intuition of finding paths from the unknown variable to the grounded entities, we have developed a path-based approach depicted in Algorithm 1 that shows the steps for transforming the AMR Graph $\mathcal{G}$ into Query Graph $\mathcal{Q}$. As \textit{amr-unknown} is the unknown variable in the AMR Graph, we retrieve all shortest paths (line 11 in Algorithm 1) between the \textit{amr-unknown} node and the nodes $\mathcal{V}_E$ of the AMR Graph $\mathcal{G}$ that have mappings to the KG entity set.  
%We opted for paths because of the characteristic of SPARQL that 
%The path-based approach facilitates to structurally map the predicates to the relationships in KG (detailed in next section). 
 Figure~\ref{fig:arch} shows an example of both the AMR and query graph for the question ``Which  actors starred in Spanish movies produced by Benicio del Toro?"
%We introduce an algorithm to do this transformation, and illustrate this procedure on our running example. 
Selecting the shortest paths reduces the n-ary predicates of AMR graph to only the relevant binary edges. For instance, the edge \textit{(act-01, arg0, person)} in the AMR graph in Figure~\ref{fig:arch} will be ignored because it is not in the path between \textit{amr-unknown} and any of the entities \textit{dbr:Spain} and \textit{dbr:Benicio\_del\_Toro}. 

%This way, the structure and the granularity of the AMR graph $\mathcal{G}$ would match that of $\mathcal{Q}$. 

\begin{figure}
  \centering
    \includegraphics[width=0.5\textwidth]{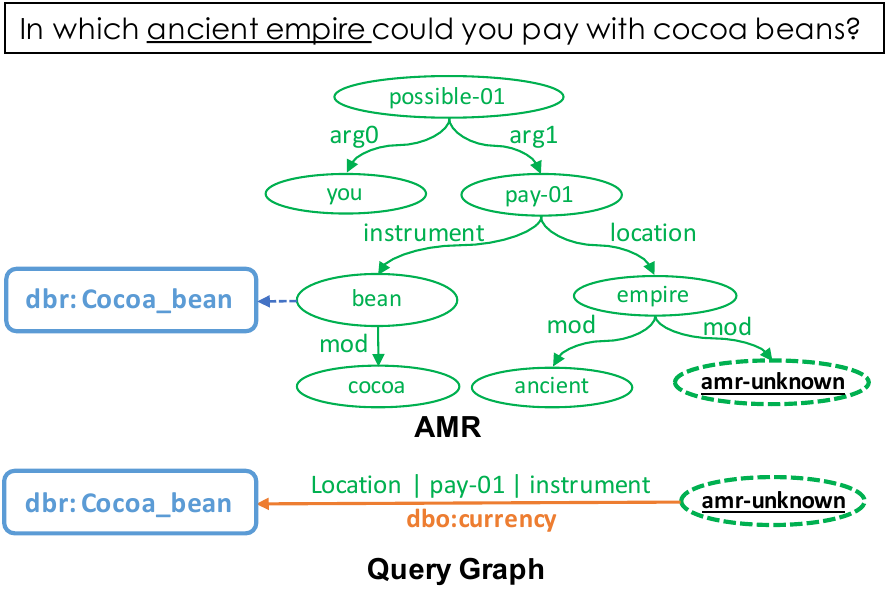}
    \caption{An example of granularity and structural mismatch between AMR and query graphs of the question \textit{In which ancient empire could you pay with cocoa beans?}. The binary relation \textit{dbo:currency} corresponds to the combination of two edges (\textit{location}, \textit{instrument}) and one node (\textit{pay-01}) in the AMR graph. } 
  \label{fig:cocoa_bean}
\end{figure}

%While the paths retrieved from AMR at line 11 in Algorithm 1 addresses the n-ary predicates challenge, they still have the structural mismatch and granularity mismatch to that of KG triples. 
%represents AMR's where nodes are predicates and concepts and edge-labels are \textcolor{red}{Pavan:add}. In order to transform this structure to a structure that can be aligned to KG (SPARQL sub-graph), where nodes are entities or variables and edge-labels are predicates we perform Steps 13-23 which performs the following: 
Structural and granularity mismatch between the AMR and query graph occurs when multiple nodes and edges in the AMR graph represent a single relationship in the query graph. This is shown in Figure~\ref{fig:cocoa_bean}. The path consists of one AMR node and 2 edges between \textit{amr-unknown} and \textit{cocoa bean}:  (\textit{amr-unknown},  location,  \textit{pay-01},  instrument,  \textit{cocoa-bean}).\footnote{For the purposes of path generation, the nodes \textit{amr-unknown} and \textit{empire} are considered equivalent because the \textit{mod} edge is a descriptor in AMR (line 8 in Algorithm 1)} In such cases, we collapse all nodes that represent predicates (like \textit{pay-01}, \textit{star-01}, etc.) into an edge, and combine it with surrounding edge labels, giving  (\textit{location} $|$ \textit{pay-01} $|$ \textit{instrument}). This is done by line 18 of Algorithm 1 where the eventual query graph $\mathcal{Q}$ will have one edge with merged predicated from AMR graph $\mathcal{G}$ between the non-predicates ($\mathcal{A_C}$).

% \begin{itemize}
%     \item Propagating the amr-unknown to its parent when it's a modifier or has a domain relationship with the parent node (Step X-Y in Algorithm 1). In Example 1 and Example 2 of Figure~\ref{fig:arch}, the amr-unknown is propagated to its parent node, \texttt{person} and \texttt{empire}.
%     \item Due to the granularity mismatch, the AMR graph can have multiple consecutive predicates in the paths retrieved. 
%     %For example the path from amr-unknown to the resolved entity (cocoa-bean) in the AMR parse of 'In which ancient empire could you pay with cocoa beans?' (shown above) 
    
%     In Example 2 of Figure~\ref{fig:arch}, the path consists of three predicates between \texttt{amr-unknown} and \texttt{cocoa bean};  (\textbf{amr-unknown}\footnote{nodes are indicated by boldface, and the rest are edges},  location,  pay-01,  instrument,  \textbf{cocoa-bean}). In such cases, we collapse predicates that occur without any concept in between. The predicates are gathered into a single edge-label  (location|pay-01|instrument). The \textit{collapse function} is called at line \textcolor{red}{X} of Algorithm 1. 
    
%     %\textcolor{red}{Pavan: Rewrite the example properly.} 
%     %This results in the semantic path below:
%     %    \texttt{e-empire $\rightarrow$ (location|pay-01|instrument) $\rightarrow$ b-cocoa bean}
% \end{itemize}

Returning to the example in Figure~\ref{fig:arch}, Algorithm 1 (line 25) outputs the query graph $\mathcal{Q}$ with the following two paths, which bear structural similarity to the knowledge graph:

{\small
\begin{itemize}\setlength\itemsep{-0.1em}
\item \texttt{amr-unknown $\rightarrow$ star-01 $\rightarrow$ movie $\rightarrow$ \\ country $\rightarrow$ Spain}
\item \texttt{amr-unknown $\rightarrow$ star-01 $\rightarrow$ movie $\rightarrow$ produce-01 $\rightarrow$ Benicio del Toro}
\end{itemize}
}

Note that in the above paths, edge labels reflect the predicates from the AMR graph (\textit{star-01}, \textit{produce-01}, and \textit{mod}). Our next step is to resolve these edge labels to its corresponding relationships from the underlying KG. To do so, we perform relation linking as described below.
\begin{algorithm}[t]
\small
% \SetKwFunction{buildRelation}{buildRelation}
\SetAlgoLined
\nl\textbf{Input} : Question text $t$, AMR graph $\mathcal{G : \langle V_G, E_G \rangle}$ having a set of nodes $\mathcal{V}_e \in \mathcal{V_G}$, each linked to a named entity in the KG \\
\nl\textbf{Returns} : Query graph $\mathcal{Q : \langle V_Q, E_Q \rangle}$ \\
 \vspace{1mm}
 Initialize $\mathcal{Q : \langle V_Q, E_Q \rangle, V_Q := \{\}, E_Q := \{\}}$ \\
 \vspace{2mm}

 \nl\uIf{$t$ is imperative}{
    \nl let source node (imperative predicate) be $r$ \\
    \nl set $q$ as \textit{amr-unknown} where $edge(r, q, \textit{`arg1'}) \in \mathcal{G_E} $ \\
    \nl delete $r$ and its edges from $\mathcal{G}$ \\
  }
%   \nl \uElseIf{$q_t$ is interrogative}{
%     \nl let source node (imperative predicate) be $r$ \\
%     \nl set $q$ as amr-unknown where $(r, arg0, q) \in \mathcal{G} $ \\
%   }
  \nl $a := $ \textit{amr-unknown} node\\
  \nl \uIf{$\exists~ b : edge(a, b, \textit{`mod'}) \in \mathcal{E_G}$}{
    \nl $a := b$\\
  }

 \vspace{2mm}
 
 \nl \For{$e$ in $\mathcal{V}_e$}
 {
    \nl $amrPath := \textbf{getShortestPath}(\mathcal{G}, a, e)$ \label{alg:getPath} \\ 
    \nl let $amrPath$ be $[a, n_1, n_2, ..., n_k]$, where $n_k = e$\\
    \nl $collapsedPath := [a]$ \\
    \nl $ n' := a $ \\
    \nl $ relBuilder := $ ` ' \\
    \nl \For{$i$ : $1 \rightarrow k$}
    {
     \nl \uIf{$n_i \in \mathcal{A_P}$}{
     \nl $ relBuilder := $\\
        $relBuilder + \textbf{getRel}_G(n', n_i) $ \\
     }
     \nl \uElseIf{$n_i \in \mathcal{A_C}$} {
     \nl append $n_i$ to $collapsedPath$  \\
     \nl add node $n'$ to $\mathcal{V_Q}$ \\
     \nl add $edge(n', n_i, relBuilder)$ to $\mathcal{E_Q}$ \\ 
     \nl $n' := n_i$ \\ 
     \nl $relBuilder := $ ` ' \\
     }
    }
    
 }
 \nl $\mathcal{Q} := doRelLinking(\mathcal{Q})$ \\
% \vspace{3mm}
\nl  \textbf{return} $\mathcal{Q}$ 
 \caption{AMR to triples}
 \label{algo:path_based}
\end{algorithm}

\noindent \textbf{Relationship Linking.}
%The task of the \textit{relation extraction and linking} (REL) module is to enrich the AMR representation by linking the frames and edges in the AMR graph to their corresponding (semantically equivalent) relations in the KB. {\sysname} uses the state-of-the-art REL system SLING~\cite{SLING}. 
%The relation linking function maps the AMR predicates in the query graph to their corresponding (semantically equivalent) relations in the KB. 
{\sysname} uses SemREL~\cite{semREL}, a state-of-the-art relation linking system that takes in the question text and AMR predicate as input and returns a ranked list of KG relationships for each triple. The cartesian product of this
represents a ranked list of candidate query graphs, and we choose the highest-ranked \textit{valid} query graph (a KG subgraph with unbound variables). 
% ***Q: what does valid mean? does it try running the query against the KB or is this domain/range checking?
As shown in Figure~\ref{fig:arch}, the output of this module produces query graph $\mathcal{Q}$ with \textit{star-01} and \textit{produce-01} mapped to DBpedia relations \textit{dbo:starring} and \textit{dbo:producer}. This will be the \where clause of the final \sparql query.

\subsubsection{Logic Generation}
\label{sec:logic}
Our query graph can be directly translated to the \where clause of the \sparql. We use existential first order logic (FOL) as an intermediate representation, where the non-logical symbols consist of the binary relations and entities in the KB as well as some additional functions to represent \sparql query constructs (e.g. \countsparql). 
We use existential FOL instead of directly translating to \sparql because: (a) it enables the use of any FOL reasoner which we demonstrate in our next Section~\ref{sec:reasoner}; (b) it is compatible with reasoning techniques beyond the scope of typical KBQA, such as temporal and spatial reasoning; (c) it can also be used as a step towards query embedding approaches that can handle incompleteness of knowledge graphs~\cite{ren2020beta,cohen2020scalable,sun2020faithful}. The Query Graph from Section~\ref{sec:approach} can be written as a conjunction in existential first order logic as shown in Figure~\ref{fig:arch}.
% We developed a logical formalism for this task to create a bridge between AMR and SPARQL that can represent both declarative and procedural knowledge. This formalism supports binary relations, which are ubiquitous in linked open data, and higher-order functional predicates to support aggregation and manipulation of sets of variable bindings. Importantly, this formalism is not restricted to the SPARQL query language. 
% %It follows the same syntactic conventions as the OpenCyc and NextKB projects,\footnote{\url{www.qrg.northwestern.edu/nextkb/}} enabling support for a broad range of semantics, from standard logical operators and the use of independent reasoners (see next Section) to DBpedia predicates and functions that emulate SPARQL constructs. 
% We developed a rule-based approach that transforms the $\mathcal{Q}$ to this logical formalism. This module consists of the following rules:
 %We want to represent the intermediate logic form as Existential First Order Logic~\cite{betaembeddings, query2box} with predicates from the corresponding KG. This allows us for better use of an out of the box logical reasoner. .
 
The current logic form supports \sparql constructs such as \select, \countsparql, \ask, and \sortsparql which are reflected in the types of questions that our system is able to answer in Table~\ref{tab:supported_types}. The heuristics to determine these constructs from AMR are as follows:

\noindent \textbf{Query Type:} This rule determines if the query will use the \ask or \select construct. Boolean questions will have AMR parses that either have no \textit{amr-unknown} variable or have an \textit{amr-unknown} variable connected to a \textit{:polarity} edge (indicating a true/false question). In such cases, the rule returns \ask, otherwise it returns \sparql.

\noindent \textbf{Target Variable:} This rule determines what unbound variable follows a \sparql statement. As mentioned in Section \ref{sec:approach}, the \textit{amr-unknown} node represents the missing concept in a question, so it is used as the target variable for the query. The one exception is for questions that have an AMR predicate that is marked as imperative, e.g. in Figure \ref{fig:amr_examples} (center) a question beginning with ``Count the awards ..." will have \textit{count-01} marked as imperative. In these cases, the algorithm uses the \textit{arg1} of the imperative predicate as the target variable (see Algorithm \ref {algo:path_based}, line 3).
% the top level AMR frame would represent the request/command of the question, and the object role (\texttt{ARG1}) typically represents the entity being requested. 

\noindent \textbf{Sorting:} 
This rule detects the need for sorting by the presence of superlatives and quantities in the query graph prior to relation linking. Superlatives are parsed into AMR with \textit{most} and \textit{least} nodes and quantities are indicated by the PropBank frame \textit{have-degree-91}, whose arguments determine: (1) which variable in $\mathcal{V}$ represents the quantity of interest, and (2) the direction of the sort (ascending or descending).   
%Questions that require sorting can be detected by the presence of \textit{have-degree-91} propbank frame. The AMR to logic translator must determine the AMR variable that corresponds to the quantity of interest and sorting direction (ascending for ~\textit{least} and descending for ~\textit{most}).
% \item {\bf $V$-Rule:} Resolving KB entity URIs to AMR variables,

\noindent \textbf{Counting:} 
%Identifying queries that require counting the bindings of the target variable in $\mathcal{V}$. %We developed a neuro-symbolic approach to detect questions that require counting operations on the bindings of the identified target variable. We first employ a neural language model based service to predict the answer type of the given question. If the predicted answer type is \textit{CARDINAL}, then we find the KB type of the unknown variable by looking at the domain or range of the triple in which it appears.
% We train a BERT-based classification model on the data from the SeMantic AnsweR Type (SMART) prediction task\textcolor{red}{CITE} 
This rule determines if the \countsparql aggregation function is needed by looking for PropBank frame \textit{count-01} or AMR edge \textit{:quant} connected to \textit{amr-unknown}, indicating that the question seeks a numeric answer.
However, questions such as ``How many people live in London?" can have \textit{:quant} associated to \textit{amr-unknown} even though the correct query will use \textit{dbo:population} to directly retrieve the numeric answer without the \countsparql aggregation function. We therefore exclude the \countsparql aggregation function if the KB relation corresponding to \textit{:quant} or \textit{count-01} has a numeric type as its range. 

\subsection{Reasoner}
\label{sec:reasoner}

%Given the logic forms from AMR to Logic, we equip {\sysname} with a neuro-symbolic reasoner that has  access to the KB to provide the answer. 

% Reasoning in KBQA systems has mostly been handled by tightly integrating rules as a part of the overall system~\cite{diefenbach2020towards,zou2014natural,vakulenko2019message}. 
With the motivation of utilizing modular, generic systems, \sysname~uses a First Order Logic, neuro-symbolic reasoner called Logical Neural Networks (LNN)~\cite{riegel2020logical}. %\sysname~is structured in a such a manner to facilitate easy incorporation of increasingly capable reasoners, as demanded by the complexity of questions. 
% LNN is a  neural network architecture in which neurons model a rigorously defined notion of weighted fuzzy or classical first-order logic. LNN takes as input the logic forms from AMR to Logic and has access to the KB to provide an answer. 
%We equip \sysname~with such a neuro-symbolic reasoner to facilitate expressive reasoning including handling inconsistencies and incompleteness of the knowledge base. 
% On the other hand,  
%Given the logic forms from AMR to Logic, we equip our system with a neuro-symbolic reasoner, Logical Neural Network (LNN)~\cite{riegel2020logical} that has  access to the KB to provide the answer. 
% Based on the KBQA datasets and the expressiveness of the associated DBpedia ontologies 
This module currently supports two types of reasoning: type-based, and geographic. 
Type-based reasoning is used to eliminate queries based on inconsistencies with the type hierarchy in the KB.
% For instance, ``Does Neymar play for Real Madrid" produced  \textit{starring(Neymar, Real\_Madrid\_C)} with faulty relation linking. \textit{starring} relationship in DBpedia ontology has \textit{dbo:Person} and \textit{dbo:Film} as its domain and range. Since the produced triple does not conform to the ontology the reasoner rejects the query graph. 
On the other hand, a question like ``Was Natalie Portman born in United States?" requires geographic reasoning because the entities related to \textit{dbo:birthPlace} are generally cities, but the question requires a comparison of countries. 
% {\color{red}
This is addressed by manually adding logical axioms to perform the required transitive reasoning for property \textit{dbo:birthPlace}.
% } 
% Both the type-based and geographic reasoning is a demonstration on the use of out-of-the-box logical reasoners and human intervention for handling complex questions. 
% {\color{blue}
We wish to emphasize that the intermediate logic and reasoning module allow for \sysname~to be extended for such complex reasoning in future work. 

\section{Experimental Evaluation}
\label{sec:evaluation}

The goal of the work is to show the value of AMR as a generic semantic parser on a modular KBQA system. In order to evaluate this, we first perform an end-to-end evaluation of {\sysname} (Section~\ref{sec:end-end-evaluation}). Next, we discuss some qualitative and quantitative results on the value of AMR for different aspects of our KBQA system (Section~\ref{sec:qualitative_results}). Finally, in support of our modular architecture, we evaluate the individual modules that are used in comparison to other state of the art approaches (Section~\ref{sec:individual_modules}).  

\subsection{Datasets and Metrics}
% \textbf{Datasets: }
To evaluate {\sysname}, we used two standard KBQA datasets on DBpedia. 
%We narrowed down to DBpedia datasets due to the availability of an ontology that can be extended for reasoning using LNN in the future~\cite{topper2012dbpedia,hohenecker2020ontology}.\footnote{Wikidata also has an ontology, however existing KBQA datasets on Wikidata~\cite{lcquad2} do not have good enough baselines to be compared.}

\noindent - \textbf{QALD - 9}~\cite{qald7} dataset has 408 training and 150 test questions in natural language, from DBpedia version 2016-10. Each question has an associated \sparql query and gold answer set. Table~\ref{tab:supported_types} shows examples of all the question types in the QALD dev set.

% LC-QuAD, on DBpedia version 2016-04, 
\noindent - \textbf{LC-QuAD 1.0}~\cite{lcquad} is a dataset with 5,000 questions based on templates and more than 80\% of its questions contains two or more relations. Our modules are evaluated against a random sample of 200 questions from the training set. LC-QuAD 1.0 predominantly focuses on the multi-relational questions, aggregation (e.g. {\countsparql}) and simple questions from Table~\ref{tab:supported_types}.

%We evaluated the whole pipeline on QALD test set (150 questions) and LC-QuAD test set (1,000 questions). 
\noindent - \textbf{Dev Set.} We also created a randomly chosen development set of 98 QALD-9 and 200 LC-QuAD 1.0 questions for evaluating individual modules. 

\noindent - \textbf{Metrics.} We report  performance based on standard precision, recall and F-score metrics for the KBQA system and other modules. For the AMR parser we use the standard Smatch metric~\cite{cai2013smatch}. 

\begin{table}
\small
\centering
\begin{tabular}{lllll}
\hline
            &  Dataset  & P & R & F  \\
\hline
WDAqua  & QALD-9  & 26.09   &   26.7 & 24.99  \\
gAnswer     & QALD-9  & 29.34 & \textbf{32.68} & 29.81               \\
% \cite{vollmers2020knowledge} & QALD-9 & & & \\
{\sysname}  & QALD-9   & \textbf{31.89} & 32.05 & \bf 31.26   \\
% {\sysname}  & QALD-9   & 31.41 & 32.16 & \bf 30.88 \\
\midrule
WDAqua  & LC-QuAD 1.0  & 22.00   &   38.00 & 28.00  \\
QAMP       & LC-QuAD 1.0  &  25.00 & \textbf{50.00} & 33.00            \\
% \cite{liang2021querying} & LC-QuAD & 1.00 & 29.20  & 29.20\\
% \cite{vollmers2020knowledge} & LC-QuAD & & & \\
{\sysname}     & LC-QuAD 1.0  & \textbf{44.76} & 45.82 &  \bf44.45  \\
% {\sysname}     & LC-QuAD  & 38.19 & 40.39 & \bf38.29 \\
\hline
\end{tabular}
\caption{{\sysname} performance on QALD-9 and LC-QuAD 1.0}
\label{tab:pipeline_performance}
\end{table}

\subsection{End-to-end Evaluation}
\label{sec:end-end-evaluation}
\noindent \textbf{Baselines:} 
We evaluate \sysname~against four systems: GAnswer~\cite{zou2014natural}, QAmp~\cite{vakulenko2019message}, WDAqua-core1~\cite{diefenbach2020towards}, and a recent approach by ~\cite{liang2021querying}. GAnswer is a graph data-driven approach and is the state-of-the-art on the QALD dataset. QAmp is another graph-driven approach based on message passing and is the state-of-the-art on LC-QuAD 1.0 dataset. WDAqua-core1 is knowledge base agnostic approach that, to the best of our knowledge, is the only technique that has been evaluated on both QALD-9 and LC-QuAD 1.0 on different versions of DBpedia. Lastly, Liang et al. ~\cite{liang2021querying} is a recent approach that uses an ensemble of entity and relation linking modules and train a Tree-LSTM model for query ranking. 
%To the best of our knowledge, WDAqua-core1 is the only technique that has evaluated on both QALD-9 and LC-QuAD on different versions of DBpedia.

\noindent \textbf{Results:} Table~\ref{tab:pipeline_performance} shows the performance of \sysname~compared to state-of-the-art approaches on QALD and LC-QuAD 1.0 datasets.
On QALD-9 and LC-QuAD 1.0, \sysname~achieves state-of-the-art performance. It outperforms WDAqua and gAnswer on QALD-9. 
% significantly and achieves a better performance compared to   by around 1.45 percentage points on F1. 
% On LC-QuAD, however, \sysname~ achieves similar performance to the state-of-the-art approach; QAmp. 
Furthermore, {\sysname}'s performance on LC-QuAD 1.0 significantly outperforms QAmp by 11.45 percentage points on F1. 
%Liang et al. ~\cite{liang2021querying} report an F1 score of 68\% on a subset of 2,430 question from LC-QuAD train and test. This subset consists only of the questions they could solve. 

Due to difference in evaluation setup in~\citet{liang2021querying}, we reevaluated their system on the same setup and metrics as the above systems. Given the test set and the evaluation, ~\cite{liang2021querying}'s F1 score reduces to 29.2\%\footnote{~\citet{liang2021querying} report an F1 score of 68\% on a different subset of LC-QuAD 1.0. They also consider only questions that returns an answer which is a different setup from the rest of the systems.}. We exclude this work from our comparison due to lack of standard evaluation. 

% \textcolor{red}{Unsupported by eval}
% Approaches like QAmp and WDAqua-core1 attempt to find the semantic meaning of the question directly over the KG whereas gAnswer try to infer a form of semantic parse using rules and modifications on the dependancy parse. This approach falls short of a dedicated semantic parser; e.g. AMR. 
% By utilizing AMR to get the semantic representation of the question with less ambiguity, \sysname~is able to generalize to sentence structures that come from very different distributions, and achieve state-of-the-art performance on both QALD-9 and LC-QuAD. 

%----------------------------------------------------------------------------------%
\begin{table}
\centering
\small
%http://gerbil-qa.aksw.org/gerbil/experiment?id=201810060001
%https://arxiv.org/pdf/1908.06917.pdf
\resizebox{\linewidth}{!}{
\begin{tabular}{lccc}
\hline
                   &AMR3.0 &  QALD-9  & LC-QuAD 1.0\\
\hline
% SPRING             & 84.5   & -      & -  \\
stack-Transformer  & 80.00   & 87.91  &  84.03\\
\hline
\end{tabular}
}
% \caption{AMR parsing performance (Smatch) against SOTA~\cite{bevilacqua2021one} on the AMR2.0 test and QALD-9, LC-QuAD dev sets.}
\caption{AMR parsing performance (Smatch) on the AMR3.0 test and QALD-9, LC-QuAD 1.0 dev sets.}
\label{tab:amrresult}
\end{table}

\subsection{Performance Analysis of AMR}  
\label{sec:qualitative_results}
\noindent\textbf{AMR Parsing.}
%As AMR parser, we use the transition-based approach developed for stack-LSTMs \cite{ballesteros2017amr}. We incorporate the recent oracle improvements and the hard-attention stack-Transformer parser from \cite{anon2020a} using the RoBERTa large with average of layers from \cite{anon2020b}. This results in a high performance parser yielding state-of-the-art performance. 
We manually created AMRs for the train and dev sets of QALD and LC-QuAD 1.0 questions. The performance of our stack-transformer parser on both of these datasets is shown in Table~\ref{tab:amrresult}. 
%\textcolor{red}{Write about Smatch not capturing the paths performance well enough.} 
%To further improve performance on QALD and LD-QUAD corpora, a mixed-training plus fine-tuning %strategy was followed. 
The parser is trained on the combination of human annotated treebanks and a synthetic AMR corpus. Human annotated treebanks include AMR3.0 and 877 questions sentences (250 QALD train + 627 LC-QuAD 1.0 train sentences) annotated in-house. The synthetic AMR corpus includes ~27k sentences obtained by parsing LC-QuAD 1.0 and LC-QuAD 2.0~\cite{jens-lehmann2019} training sentences, along the lines of ~\cite{youngsuklee-etal-emnlp2020}. 
% The results for this configuration as well as the most recent SOTA 
% \cite{bevilacqua2021one} is displayed in Table \ref{tab:amrresult}.
%\footnote{Downloadable from https://www.computing.dcu.ie/~jjudge/qtreebank/}

%----------------------------------------------------------------------------------%

\begin{figure*}[h!]
  \centering
    \includegraphics[width=1\textwidth]{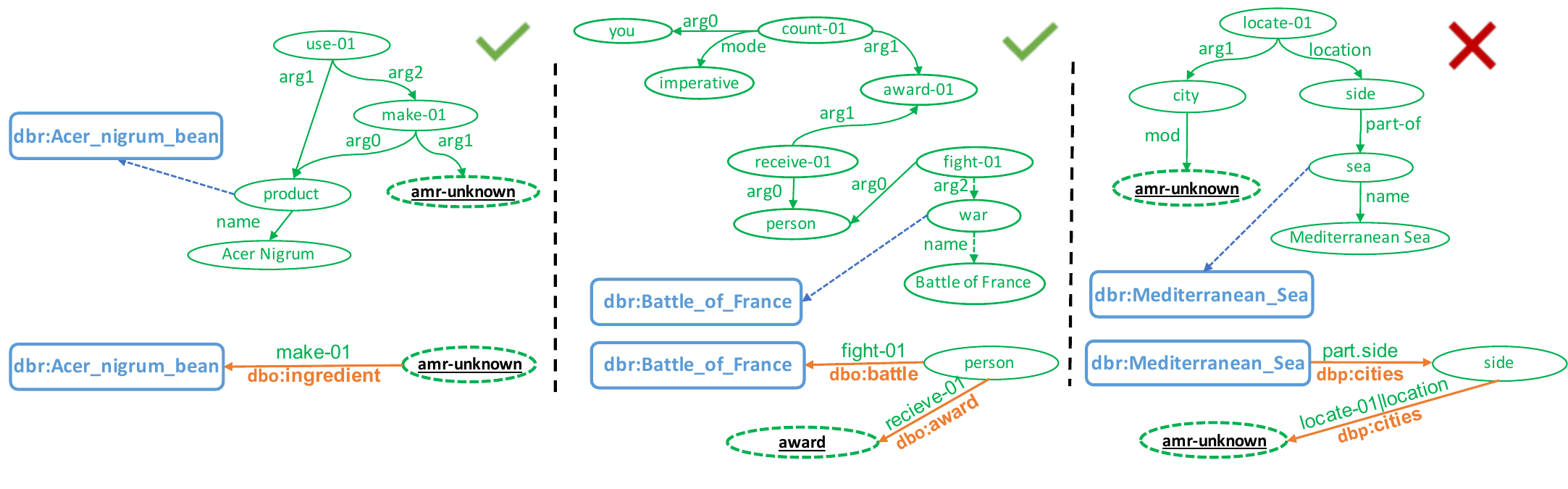}
    \caption{AMR and query graphs for the questions \textit{``Acer nigrum is used in making what?"}, \textit{``Count the awards received by the ones who fought the battle of france?"} and  \textit{``What cities are located on the sides of mediterranean sea?"} from LC-QuAD 1.0 dev set} 
  \label{fig:amr_examples}
\end{figure*}

% \begin{table}
% \centering
% \small
% \begin{tabular}{llll}
% \toprule
% Query Feature & Correct & Total & Correct (\%)  \\
% \midrule
% SELECT        & 156   & 186     & 83.9         \\
% ASK           & 14    & 14      & 100.0           \\
% COUNT         & 25    & 31      & 80.6         \\
% \midrule
% 1-Hop         & 43    & 63      & 68.2         \\
% 2-Hop         & 86   & 137      & 62.8         \\
% \bottomrule
% \end{tabular}
% \caption{Query constructs prediction (LC-QuAD dev)}
% \label{tab:structure}
% \end{table}

\begin{table}
\centering
\small
\begin{tabular}{llll}
\toprule
Query Feature & Correct & Total & Correct (\%)  \\
\midrule
SELECT        & 164   & 186     & 88.2         \\
ASK           & 14    & 9       & 64.3           \\
COUNT         & 25    & 31      & 80.6         \\
\midrule
1-Hop         & 50    & 63      & 79.4         \\
2-Hop         & 96    & 137     & 70.1         \\
\bottomrule
\end{tabular}
\caption{Query constructs prediction (LC-QuAD 1.0 dev)}
\label{tab:structure}
\end{table}

\noindent \textbf{AMR-based Query Structure}
%SLING "Macro F1 score": "0.4053588144935394",
% Number of ask questions:  14 , correct 14 (percentage: 1.0)
% Number of select questions:  186 , correct 156 (percentage: 0.8387096774193549)
% Number of count questions:  31 , correct 25 (percentage: 0.8064516129032258)
% Number of 1 triples questions:  63 , correct 43 (percentage: 0.6825396825396826)
% Number of 2 triples questions:  137 , correct 86 (percentage: 0.6277372262773723)
%SimREL: "Macro F1 score": "0.5176204417551666",
% Number of ask questions:  14 , correct 9 (percentage: 0.6428571428571429)
% Number of select questions:  186 , correct 164 (percentage: 0.8817204301075269)
% Number of count questions:  31 , correct 25 (percentage: 0.8064516129032258)
% Number of 1 triples questions:  63 , correct 50 (percentage: 0.7936507936507936)
% Number of 2 triples questions:  137 , correct 96 (percentage: 0.7007299270072993)
{\sysname} leverages many of the AMR features to decide on the correct query structure. As shown in Section~\ref{sec:logic}, {\sysname} relies on the existence of certain PropBank predicates in the AMR parse such as \textit{have-degree-91}, \textit{count-01}, \textit{amr-unknown} to decide on which \sparql constructs to add. In addition, the AMR parse determines the structure of the \where clause. In Table~\ref{tab:structure}, we show the accuracy of each one of these rules on LC-QuAD 1.0 dev dataset. 
% Overall, {\sysname} identified all ASK (boolean) questions correctly and achieved more than 80\% accuracy for COUNT and SELECT questions. Using AMR and the path-based approach, {\sysname} was able to correctly predict the total number of constraints with comparable accuracies of 68\% and 62\% for single and two-hops, respectively.  \sysname~finds the correct query structure for complex questions almost as often as for simple questions, completely independent of the KG.  
Overall, {\sysname} identified 64\% of \ask (boolean) questions correctly and achieved more than 80\% accuracy for \countsparql and \select questions. Using AMR and the path-based approach, {\sysname} was able to correctly predict the total number of constraints with comparable accuracies of 79\% and 70\% for single and two-hops, respectively.  \sysname~finds the correct query structure for complex questions almost as often as for simple questions, completely independent of the KG.  
% {\color{red} Add a sentence to conclude}

%----------------------------------------------------------------------------------%

Figure~\ref{fig:amr_examples} shows two examples illustrating how AMR lends itself to an intuitive transformation to the correct query graph, as well as a third example where we fail. Here the AMR semantic parse can not be matched to the underlying KG, since `side' is an extra intermediate variable that leads to an additional constraint in the query graph.
{
% Figure~\ref{fig:amr_examples} shows three examples from LC-QuAD dev set. The first question \textit{``Acer nigrum is used in making what?"} is a single hop example that requires collapsing the two PropBank predicates \textit{use-01} and \textit{make-01} between the \textit{amr-unknown} and the entity. The second question \textit{``Count the awards received by the ones who fought the battle of france?"} is a two-hop example with multiple variables which the AMR graph captures well. It is worth noting here the lexical gap in the relations of this example where \textit{fight-01} should be mapped to \textit{dbo:battle} and \textit{receive-01} corresponds to \textit{dbo:award} which makes relation linking is this setting very challenging. Finally, the third question \textit{``what cities are located on the sides of mediterranean sea?"} shows an example where the AMR graph, while being correct, can not be matched as is to the underlying KG. The reason is that \textit{side} node is a concept and could not be collapsed with our path-based approach which resulted in two triples in the query graph instead of one.  
}

%----------------------------------------------------------------------------------%

\begin{table*}
\centering
\small
\begin{tabular}{l|l|c}
\toprule
Question Type/Reasoning & Example & Supported \\ \midrule
Simple                  & Who is the mayor of Paris                                          & \checkmark                             \\
Multi-relational  & Give me all actors starring in movies directed by William Shatner. & \checkmark                             \\
Count-based             & How many theories did Albert Einstein come up with?                & \checkmark                             \\
Superlative             & What is the highest mountain in Italy?                             & \checkmark                             \\
Comparative             & Does Breaking Bad have more episodes than Game of Thrones?         & \multicolumn{1}{l}{}          \\
Geographic              & Was Natalie Portman born in the United States?                     & \checkmark                             \\
Temporal                & When will start $[sic]$ the final match of the football world cup 2018?    & \multicolumn{1}{l}{}          \\ \bottomrule
\end{tabular}
\caption{Question types supported by \sysname~, with examples from QALD}
\label{tab:supported_types}
\end{table*}

\noindent \textbf{Supported Question Types.} Table~\ref{tab:supported_types} shows the reasoning and question types supported by \sysname~. Our transformation algorithm applied to AMR parses supports simple, multi-relational, count-based, and superlative question types. LNN performs geographic reasoning as well as type-based reasoning to rank candidate logic forms. Addressing comparative and temporal reasoning is a part of our future work. %Our path-based approach in conjunction with AMR as the semantic parse is able to handle all of the mentioned supported types. The geographic reasoning is particularly performed by LNN which also re-ranks logic forms using its type-based reasoning capability. 

\subsection{Individual Module Evaluation}
\label{sec:individual_modules}
%In this section, we evaluate the performance of individual modules.
 %\sysname~is highly modular, to enable the replacement of any module and leverage state-of-the-art advancement in that sub-task. %Currently, all our modules exhibit SOTA performance.

\noindent \textbf{Entity and Relation Linking.}
\sysname's EL module (NMD+BLINK) consists of a BERT-based neural mention detection (NMD) network, trained on LC-QuAD 1.0 training dataset comprising of 3,651 questions with manually annotated mentions, paired with an off-the-shelf entity disambiguation model -- BLINK~\cite{Wu2019ZeroshotEL}. 
We compare the performance of NMD+BLINK approach with Falcon~\cite{sakor-etal-2019-old} in Table~\ref{tab:entity_linking_results}. NMD+BLINK performs 24\% better on F1 than Falcon (state-of-the-art) on  LC-QuAD 1.0 dev set and 3\% better on QALD-9 dev set.
% {\color{red}
Similarly, we evaluate Relation Linking on both QALD and LC-QuAD 1.0 dev sets. In particular, we used SemREL~\cite{semREL}; state-of-the-art relation linking approach which performs significantly better compared to both Falcon~\cite{sakor-etal-2019-old} and SLING~\cite{SLING} on various datasets. 
On LC-QuAD 1.0 dev, SemREL acheives F1 = 0.55 compared to 0.43 by SLING and 0.42 by Falcon.
On QALD-9, SemREL achieves 0.54 compared to 0.64 and 0.46 F1 for SLING and Falcon, respectively.
% }
%These results were used to select and improve the modules used in the overall  pipeline of~{\sysname}. 

% and DBpedia Spotlight~\cite{10.1145/2063518.2063519} with confidence threshold set to 0.5.  
% Demonstrated in Table~\ref{tab:entity_linking_results}, NMD+BLINK performs 12\% better on F1  than Falcon (best state-of-the-art) on  LC-QuAD dev set and achieves comparable performance on QALD-9 dev set.

% Demonstrated in Table~\ref{tab:entity_linking_results}, NMD+BLINK performs 30\% better on F1  than Falcon (best state-of-the-art) on  LC-QuAD dev set and 3.6\% better on QALD-9 dev set.

\begin{table}
\centering
\small
\begin{tabular}{lllll}
\hline
            &  Dataset  & P & R & F1 \\
\hline
%TagMe       & QALD-9   &           &        &         \\
% DBpedia Spotlight@0.0 & QALD-9  &    0.28       & 0.74        &  0.40       \\
% DBpedia Spotlight@0.5 & QALD-9  &    0.71       & 0.70        &  0.70       \\
Falcon     & QALD-9  &    0.81       & 0.83        &  0.82       \\
%NMD+BLINK        & QALD-9   &  0.76         &   0.88     &   \bf 0.82       \\
\textbf{NMD+BLINK}        & QALD-9   &  0.82         &   0.90     &   \bf 0.85       \\
\midrule
%TagMe       & LC-QUAD  &  0.65        &   0.77     &   0.68      \\
% DBpedia Spotlight@0.0 & LC-QuAD  &    0.17       & 0.71        &  0.28       \\
% DBpedia Spotlight@0.5 & LC-QuAD  &    0.54       & 0.66        &  0.59       \\
Falcon     & LC-QuAD 1.0  &    0.56  &   0.69   &  0.62       \\
%NMD+BLINK        & LC-QuAD  &  0.66         &  0.85      &   \bf 0.74       \\
\textbf{NMD+BLINK}        & LC-QuAD 1.0  &  0.87         &  0.86      &   \bf 0.86       \\
% \hline
% \hline
% % EARL       & QALD-9    &   0.28        &   0.35      &  0.31       \\
% Falcon     & QALD-9   &     0.45      &   0.47     &     0.46    \\
% SLING        & QALD-9    &     0.71      &      0.81  &      \bf0.76    \\
% \midrule
% % EARL       & LC-QuAD  &  0.20        &   0.39    &   0.26      \\
% Falcon     & LC-QuAD  &    0.38  &   0.47    &  0.42       \\
% SLING        & LC-QuAD  &     0.36      &     0.54   &    \bf0.43     \\

% \hline

\hline
\end{tabular}
\caption{Performance of Entity Linking modules compared to SOTA Falcon on our dev sets}
\label{tab:entity_linking_results}
\end{table}

%This combination performs the best in comparison to state-of-the-art Falcon on QALD and LC-Quad in terms of both EL module and the pipeline performance; demonstrated in Table~\ref{tab:entity_linking_results}, NLU achieves 12 F1 points higher than Falcon and on LC-QuAD dataset and comparable on QALD-9 dataset.

\noindent \textbf{Reasoner.} We investigate the effect of using LNN as a reasoner equipped with axioms for type-based and geographic reasoning.
We evaluated {\sysname}'s performance under two conditions: (a) with an LNN reasoner with intermediate logic form and (b) with a deterministic translation of query graphs to \sparql. 
On LC-QuAD 1.0 dev set, {\sysname} achieves an F1 score of 40.5 using LNN compared to 37.6 with the deterministic translation to \sparql. Based on these initial promising results, we intend to explore more uses of such reasoners for KBQA in the future.

\section{Related Work}
\label{sec:relatedwork}

Early work in KBQA  focused mainly on designing parsing algorithms and (synchronous) grammars to semantically parse input questions into KB queries~\cite{zettlemoyer2007online,berant2013semantic}, with a few exceptions from the information extraction perspective that directly rely on relation detection~\cite{yao2014information,bast2015more}.
All the above approaches train statistical machine learning models based on human-crafted features and the performance is usually limited. 

\noindent \textbf{Deep Learning Models.}  
The renaissance of neural models significantly improved the accuracy of KBQA systems~\cite{yu2017improved,wu2019general}.
Recently, the trend favors translating the question to its corresponding subgraph in the KG in an end-to-end learnable fashion, to reduce the human efforts and feature engineering.
This includes two most commonly adopted directions: (1) embedding-based approaches to make the pipeline end-to-end differentiable \cite{bordes2015large,xu2019enhancing}; (2) hard-decision approaches that generate a sequence of actions that forms the subgraph \cite{xu2018exploiting,bhutani2019}. 

On domains with complex questions, like QALD and LC-QuAD, end-to-end approaches with hard-decisions have also been developed. Some have primarily focused on generating \sparql sketches~\cite{maheshwari2019learning,chen2020formal} where they evaluate these sketches (2-hop) by providing gold entities and ignoring the evaluation of selecting target variables or other aggregation functions like sorting and counting.
%and achieved state-of-the-art or comparable results.  For example, \cite{hu2018state} build a state transition-based approach with novel actions specifically defined for KBQA.
\cite{zheng2019question} generates the question subgraph via filling the entity and relationship slots of 12 predefined question template. Their performance on these datasets show significant improvement due to the availability of these manually created  templates. Having the advantage of predefined templates does not qualify for a common ground to be compared to generic and non-template based approaches such as~\sysname, WDAqua, and QAmp.

%Approaches like~\sysname~that do not require pre-defined templates thus are more robust and can better generalize to new question types.  
%However, similar to the learning-free systems, they still require a lot of human effort on question template analysis. Compared to these approaches, {\sysname} does not require predefined question templates thus is more robust and can better generalize to new question types. {\color{red}IJCAI paper on structure generation-end2end}

%\cite{bhutani2019} is another end-to-end neural-network based graph-driven model that learns to score candidates for parts of complex questions using implicit supervision from question-answer pairs.

%Furthermore, any semantic abstractions in the aforementioned queries come directly from the KB, whereas the semantics of AMR parses in {\sysname} already capture some abstractions from language even before any KG alignment is attempted.

% \paragraph{KBQA on QALD}
\noindent \textbf{Graph Driven Approaches.} Due to the lack of enough training data for KBQA, several systems adopt a training-free approach. 
WDAqua~\cite{diefenbach2017wdaqua} uses a pure rule-based method to convert a question to its \sparql query. gAnswer~\cite{zou2014natural} uses a graph matching algorithm based on the dependency parse of question and the knowledge graph. 
%builds a semantic graph from the question and its dependency parse, then a graph matching algorithm is used to match the semantic graph to the KB to get the answer.
QAmp~\cite{vakulenko2019message} is a graph-driven approach that uses message passing over the KG subgraph containing all identified entities/relations where confidence scores get propagated to the nodes corresponding to the correct answers. 
Finally, \cite{mazzeo2016answering} achieved superior performance on QALD-5/6 with a hand-crafted automaton based on human analysis of question templates. 
A common theme of these approaches, is that the process of learning the subgraph of the question is heavily KG specific, while our approach first delegates the question  understanding to KG-independent AMR parsing. %identifies the semantic interpretation primarily using the AMR semantic parse, which is independent of the KG.

%%%%%%%%%%%%%%%%%%%%%%%%%%%%%%%%%%%%%%%%%%%%%%%%%%%%%%%%%%%%%%%%%%%%%%%%%%%%%
%%%%%%%%%%%%%%%%%%%%%%%%%%%%%%%%%%%%%%%%%%%%%%%%%%%%%%%%%%%%%%%%%%%%%%%%%%%%%
%%%%%%%%%%%%%%%%%%%%%%%%%%%%%%%%%%%%%%%%%%%%%%%%%%%%%%%%%%%%%%%%%%%%%%%%%%%%%

% In this section, we describe recent systems that perform well in the KBQA task, either on the QALD dataset or LC-QuaD v1.0. We describe three systems: gAnswer, QAmp and WSDAqua-core1.

%While most approaches have not focused on reusuability,

% Similar to \sysname, Frankenstein~\cite{singh2018frankenstein} is a system that emphasize the aspect of reusuability. Conditioned on the questions, the system learns weights for each reusuable component. A more recent work~\cite{liang2021querying} builds on Frankenstein's idea to harness its modularity. It starts by detecting the question type (e.g. ASK or SELECT) and running a phrase mapping phase that generates query candidates with entities and relations linked. It then rank these candidates using a TreeLSTM model. \cite{chen2020formal} proposed two-phased query generation approach for KBQA. 
% In the first phase, the query structure is predicted using a machine learning generative model to generate a list of candidate queries which are ranked in the second phase. 
% One primary difference to our approach is that all these methods do not support semantic parsing or use of reasoners for KBQA task. Furthermore, Frankenstein results are not competitive to the state-of-the-art approaches while \cite{chen2020formal} require end-to-end training data. 

 \noindent\textbf{Modular Approaches.} Frankenstein~\cite{singh2018frankenstein} is a system that emphasize the aspect of reusuability where the system learns weights for each reusuable component conditioned on the questions. They neither focus on any KG-independent parsing (AMR) not their results are comparable to any state of the art approaches. \cite{liang2021querying} propose a modular approach for KBQA that uses an ensemble of phrase mapping techniques and a TreeLSTM-based model for ranking query candidates which requires task specific training data. 
%\cite{chen2020formal} proposed two-phased query generation approach for KBQA.  One primary difference to our approach is that all these methods do not support semantic parsing or use of reasoners for KBQA task. 
%Furthermore,  while \cite{chen2020formal} require end-to-end training data. 
%}

\section{Discussion}
The use of semantic parses such as AMR compared to syntactic dependency parses provides a number of advantages for KBQA systems. 
First, independent advances in AMR parsing that serve many other purposes can improve the overall performance of the system. For example, on LC-QUAD-1 dev set, a 1.4\% performance improvement in AMR Smatch improved the overall system's performance by 1.2\%. Recent work also introduces multi-lingual and domain-specific (biomedical) AMR parsers, which expands the possible domains of application for this work.
Second, AMR provides a normalized form of input questions that makes {\sysname} resilient to subtle changes in input questions with the same meaning. 
Finally, AMR also transparently handles complex sentence structures such as multi-hop questions or imperative statements.
% where multiple variations are represented by the same AMR-graph. This is a significant advantage over the use of syntactic representations. 
% In addition to normalizing the predicates, AMR gives {\sysname} the structure of predicates and arguments which is evaluated based on how accurately AMR classifies questions of multiple hops. The third benefit is that the use of PropBank information in AMR has also shown to have been helpful for Relation Linking for KBQA~\cite{SLING}.

Nevertheless, the use of AMR semantic parses in {\sysname} comes with its own set of challenges: 1) Error propagation: Although AMR parsers are very performant (state-of-the-art model achieves an Smatch of over 84\%), 
inter-annotator agreement is only 83\% on newswire sentences, as noted in ~\cite{banarescu2013abstract}. 
% And our system is over-reliant on AMR paths to be correct. 
Accordingly, AMR errors can propagate in {\sysname}'s pipeline and cause errors in generating the correct answer, 2) Granularity mismatch: our proposed path-based AMR transformation is generic and not driven by any domain-specific motivation, but additional adjustments to the algorithm may be needed in new domains due to the different granularity between AMR and SPARQL 3) Optimization mismatch: Smatch, the optimization objective for AMR training, is sub-optimal for KBQA. {\sysname} requires a particular subset of paths to be correctly extracted, whereas the standard AMR metric Smatch focuses equally on all edge-node triples. We are therefore exploring alternative metrics and how to incorporate them into model training.

\section{Conclusion and Future Work}
\label{sec:discussion}
To the best of our knowledge, {\sysname} is the first system that successfully harnesses a generic semantic parser, particularly AMR, for a KBQA task. Our path-based approach to map AMR to the underlying KG such as DBpedia is first of its kind with promising results in handling compositional queries. {\sysname} is a modular system where each modules are trained separately for its own task, hence not requiring end-to-end KBQA training. %In this work, we have successfully leveraged a neuro-symbolic reasoner to perform type-based and geographic reasoning.
In future, we will explore the potential of the more expressive intermediate logic form with the neuro-symbolic reasoner for KBQA. Particularly, we intend to focus on extending {\sysname} for temporal reasoning and making it robust to handle incompleteness and inconsistencies in knowledge bases. 
%as it integrates neural networks' tolerance of noisy data and interpretable inference with hand-crafted or induced domain knowledge.  

% In comparison to graph-driven approaches that are tightly coupled to a single knowledge graph, our work can smoothly be applied to different KGs 

%---------------------------------

% \input{appendix}

\bibliographystyle{acl_natbib}
\bibliography{acl2021,ijcai21}

\end{document}

% \linenumbers
\maketitle
%--------Main paper---------------
\section{Logical Neural Networks}
% \subsection{Overview}
Logical Neural Network is a novel neural network architecture in which neurons model a rigorously defined notion of weighted fuzzy or classical first-order logic.
Arranged in a one-to-one correspondence with the operations in a system of logical formulae, LNN is capable of inference in any direction, i.e.~via normal evaluation or reverse inferences such as \emph{modus ponens}, \emph{modus tollens}, conjunction elimination, and all related inference rules. 

The LNN explicitly acknowledges the open-world hypothesis by tracking both upper and lower bounds on truth values.
This is achieved using specialized neural activation functions and computation patterns, such that proven truth value bounds propagate from each of a formula's inputs to each other input, where they are aggregated and used again in recurrent computation.
In addition, the LNN focuses computation on specific rules and groundings more likely to yield informative results---an instrumental acceleration for reasoning in first-order logic.

Constrained with sufficient penalty, a trained such model is guaranteed to converge on classical inference behavior.
With looser constraints it is able to handle incomplete formulae, minor contradictions, and other sources of uncertainty in the ground truth.
It also keeps detailed uncertainties for atoms, neurons and quantifiers in its internal logical structure, in contrast to formula-level approaches~\cite{riegel2020logical} like Markov Logic Networks~\cite{richardson2006markov}, probabilistic soft logic~\cite{bach2017hinge} or TensorLog~\cite{cohen2016tensorlog}.
\begin{algorithm}[t]
\small
% \SetKwFunction{buildRelation}{buildRelation}
\SetAlgoLined
\nl\textbf{Input} : Question logic formula $\mathcal{L}$ and support axioms $\mathcal{A}$, with predicates $\mathcal{P}$\\
\nl\textbf{Returns} : Answer entity-set $\mathcal{R}$ \\
 \vspace{3mm}
\nl Map logic formula $\mathcal{L}$ to LNN ($\mathcal{G}$) nodes/neurons $\mathcal{V}_\mathcal{L}$\\
\nl \For{$v$ in $\mathcal{V}_\mathcal{L}$}
{
    \nl let $\mathcal{L}$'s variable bindings be $B$ over preds. $\mathcal{P}_\mathcal{L}$\\
    \nl\uIf{$v$ is root($\mathcal{G}$)}{
        \nl \For{$c$ in child($v$)}
        {
            \nl recursively return SPARQL repr. of child $c$ \\
            \nl incorporate child SPARQL into $v$ SPARQL\\
        }
        \nl complete $v$ SPARQL repr.\\
        \nl execute $v$ SPARQL, obtain variable bindings $B$\\
    }
}

\nl \For{$P$ in $\mathcal{P}_\mathcal{L}$}
{
    \nl form predicate SPARQL for $P$ with bindings $B$\\
    \nl execute $P(B)$ SPARQL, obtain facts $\{P(f)\}$\\
    \nl ground LNN predicate $P$ with $\{P(f)\}$\\
}

\nl \For{$v$ in $\mathcal{V}_\mathcal{L}$}
{
    \nl\uIf{$v$ is root($\mathcal{G}$)}{
        \nl \For{$c$ in child($v$)}
        {
            \nl recursively call inference/function at child $c$ \\
        }
        \nl perform inference/function at $v$\\
    }
}

\nl  \textbf{return} $root(\mathcal{L})$ 
 \caption{Logic to grounded LNN conversion, with upward inference and query evaluation.}\label{lnn-algo}
\end{algorithm}

\subsection{LNN in \sysname}
In \sysname, LNN evaluates the logical query it receives as input from AMR to Logic to produce an answer.
We opt for LNN as it is able to mitigate issues associated with incompleteness (open world assumption) and inconsistency (arising from human errors in curation) of KBs.
The syntax tree of the first-order logic form from AMR to Logic maps bijectively to the directed graph of the LNN, which can consist of atoms/predicates, logical connectives (neurons), quantifiers and functional nodes (counting, sorting, filtering). For instance, \textit{$\arg\exists_z$(type($x$, Film) $\land$ country($x$, Spain)  $\land$ producer($x$, Benicio\_del\_Toro) $\land$ starring($x$, $z$) $\land$ type($z$, Person))} contains an existential quantifier, 2 unbound variables, and conjunctions. 

A query conjunction can weigh KB triples in each AMR to Logic hypothesis according to triple relevances given by the \textit{Relation Extractor and Linker} module. 
Simultaneous evaluation of multiple possible AMR to Logic hypotheses is also possible with a weighted disjunction using average weights of contained triples per hypothesis. 

LNN supports the open-world assumption with probabilistic semantics expressed through lower and upper bounds on truth values in the real-unit interval [0, 1], with 0 and 1 as special cases representing classical true and false.
Truth value bounds provide additional parameters in LNN and can be learnt, in order to relax facts and ontological rules from inconsistent KBs like DBpedia to avoid contradiction (lower $>$ upper bound). 
LNN has probabilistic semantics and can express uncertainty, which will be used in future work to decide between multiple entity and relation links to improve subgraph-selection.
\begin{figure}[htpb!]
  \centering
    \includegraphics[width=0.45\textwidth,trim={0.05cm 0.2cm 18.9cm 0cm},clip]{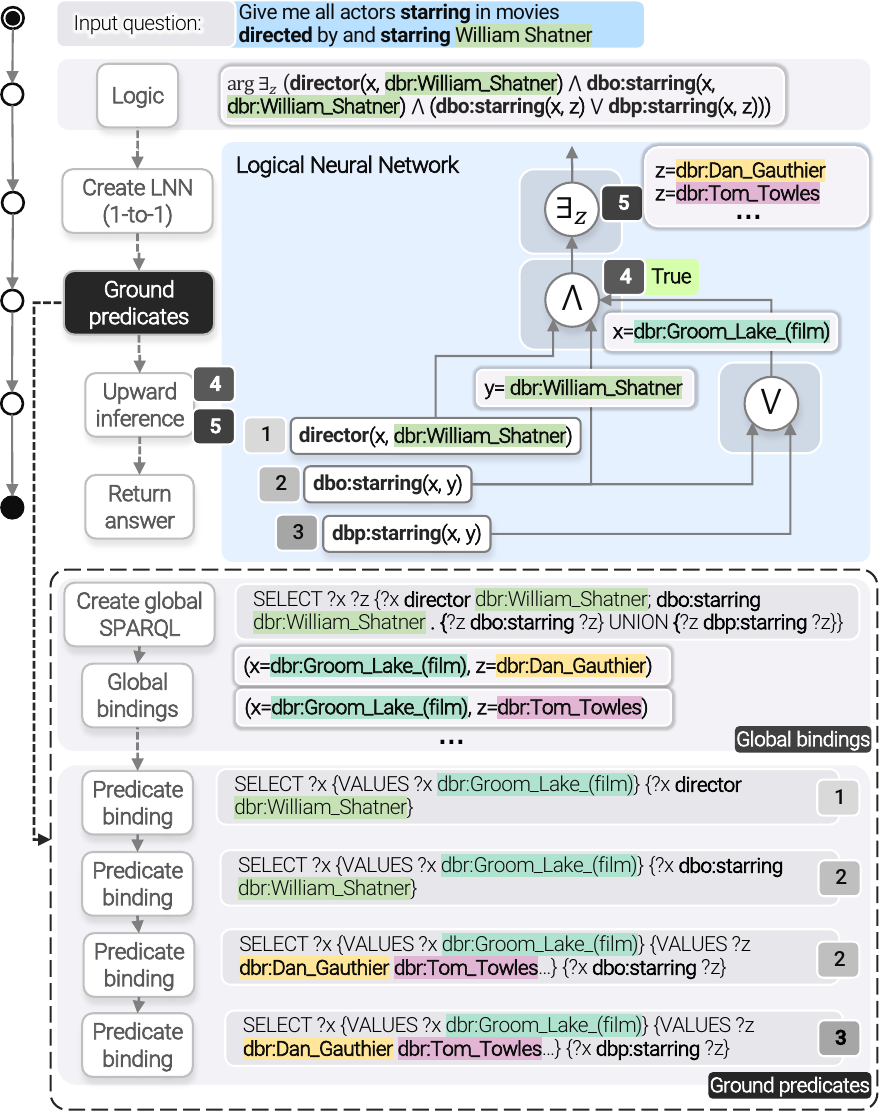}
  \caption{Multihop reasoning LNN example.} 
  \label{fig:lnn_shatner}
\end{figure}
\subsection{Grounding LNN}
Algorithm~\ref{lnn-algo} provides the description for converting Logic to LNN, grounding the LNN with external KB access, performing inference for query evaluation.
To speed computation, the LNN uses SPARQL queries to obtain data as needed from the graph database, starting from predicates with the most restrictive bindings, then using those results to bind common variables in other queried predicates.
The LNN initially forms a global SPARQL query under the provided variable bindings. This is achieved by starting from the predicate leaves, traversing to the query root node (typically an existential quantifier) and on the way, accumulating implicated groundings.

% Examples of grounding and evaluating LNN are shown in Figures~\ref{fig:lnn_del_toro}, \ref{fig:lnn_cocoa}, \ref{fig:lnn_cosmo}, \ref{fig:lnn_shatner}, \ref{fig:lnn_cuban}, \ref{fig:lnn_river}.
These then bind separate SPARQL calls that ground each predicate individually, required to determine contributions under disjunctions. Index, sort and filter functional nodes are executed by direct SPARQL inclusion, otherwise initial grounding is often excessive. Counting is procedurally performed on indicated variables for their operand groundings, although special combinations with sort and filter are omitted.

In the current QALD pipeline, the DBPedia dataset is ingested and queried using its SPARQL endpoint for generating the final answer. This also forms core knowledge store to support interfaces to various other modules in the pipeline that require access to the underlying knowledge base, notably the LNN.
\begin{figure}%[htpb!]
  \centering
    \includegraphics[width=0.45\textwidth,trim={0.05cm 0.0cm 18.9cm 0cm},clip]{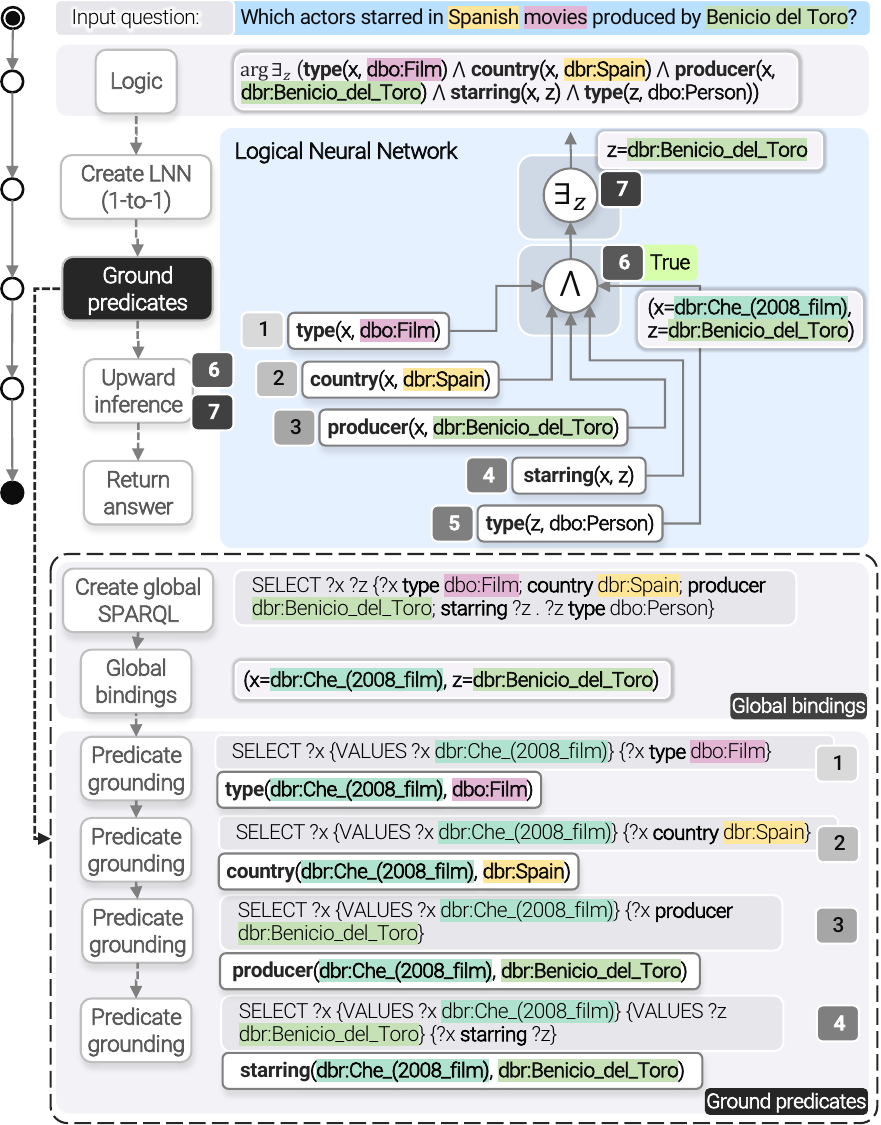}
  \caption{LNN handling of the running example ``Which actors starred in Spanish movies produced by Benicio del Toro?"} 
  \label{fig:lnn_del_toro}
\end{figure}

\begin{figure}[t!]
  \centering
    \includegraphics[width=0.45\textwidth,trim={0.05cm 2.1cm 18.9cm 0cm},clip]{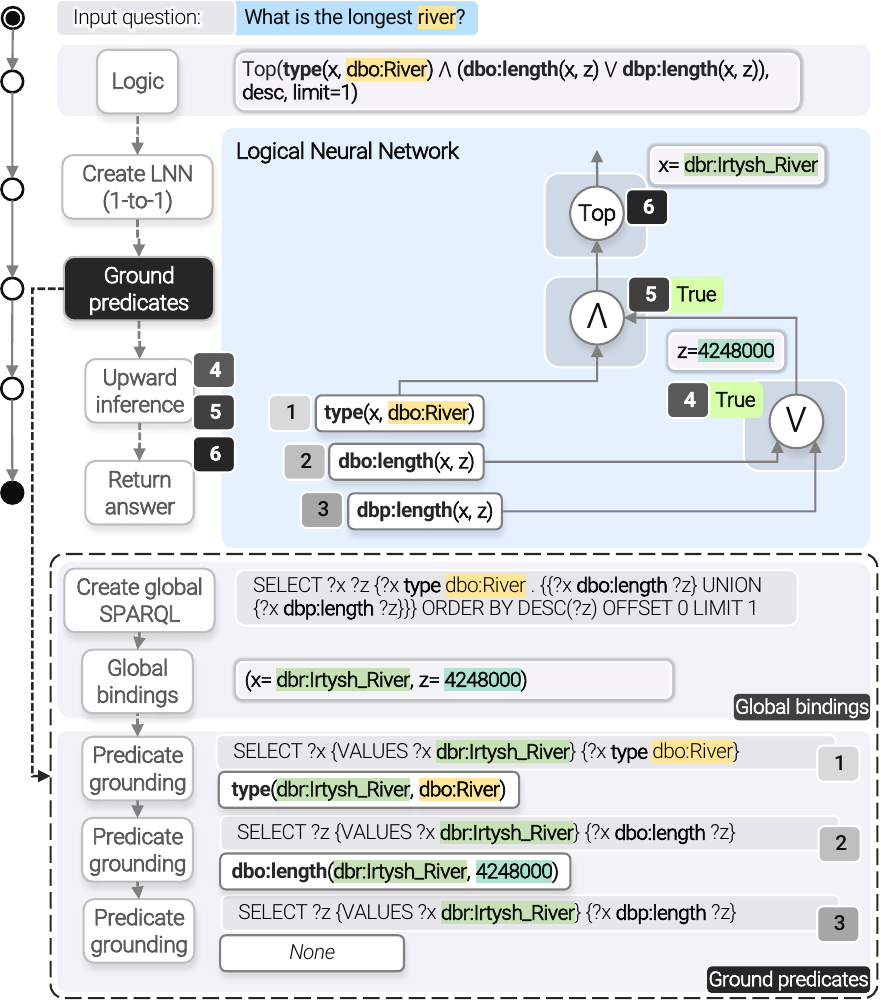}
  \caption{Superlative reasoning with LNN.} 
  \label{fig:lnn_river}
\end{figure}

\textbf{Example explanation:} Examples of grounding and evaluating LNN are shown in Figures~\ref{fig:lnn_shatner}, \ref{fig:lnn_del_toro}, and \ref{fig:lnn_river}. For the running example ``Which actors starred in Spanish movies produced by Benicio del Toro?" in Figure~\ref{fig:lnn_del_toro}, LNN performs the following steps:
% \begin{enumerate}
    \item \textbf{Mapping of Logic to LNN graph:}
    The input logic to the LNN module is $\arg\exists_z$(rdf:type($x$, dbo:Film) $\land$ dbo:country($x$, dbr:Spain) $\land$ dbo:producer($x$, dbr:Benicio\_del\_Toro) $\land$ dbo:starring($x$, $z$) $\land$ rdf:type($z$, dbo:Person)). This logic has a one-to-one mapping to the LNN graph/network, which gets a corresponding existential quantifier as root node, and a conjunction (neuron) of a set of five specifically-bound input predicates featuring in the input logic.
    
    \item \textbf{Global variable bindings:} 
    The purpose of obtaining the global variable bindings is to restrict the set of facts retrieved from the KB to only those necessary to answer the question through upward inference and computation in the LNN. For example, superlative and conditional questions using sorting and filtering typically involve evaluation over a large set of candidates, which SPARQL is relatively optimized to conduct efficiently internally. 
    
    In contrast, LNN needs to externally retrieve the underlying set of candidates to perform logical inference and various functions through computation, which incurs a retrieval delay. Unless the retrieved facts are proactively filtered to only the essential facts required to reach the correct answer, although this typically means that the answer is obtained during this step. Nevertheless, the objective here is to allow computation and inference in LNN to explicitly calculate the answer.
% \begin{figure}[b!]%[htpb!]
%   \centering
%     \includegraphics[width=0.5\textwidth,trim={0.05cm 6.6cm 18.9cm 0cm},clip]{images/lnn-cocoa.pdf}
%   \caption{LNN handling of the running example ``In which ancient empire could you pay with cocoa beans?"} 
%   \label{fig:lnn_cocoa}
% \end{figure}

    LNN performs leaves-to-root traversal of its graph of neurons and functional nodes, to construct a global SPARQL query corresponding to the root node, which is typically an existential quantifier at the outside of the logic query. This results in a SPARQL statement as follows: \texttt{SELECT ?x ?z \{?x type dbo:Film; country dbr:Spain; producer dbr:Benicio\_del\_Toro; starring ?z . ?z type dbo:Person\}}
    
    LNN performs information retrieval from the KB through calling of the SPARQL query, obtaining tuples for $(x, z)$ that subsequently apply as variable bindings in further SPARQL calls for grounding individual predicates. A variable binding means that only values observed in the obtained tuples are permitted for the corresponding variables, and that this restriction is added to fact retrieval queries that follow.
    In particular, only one tuple is obtained for the global bindings, namely (x=dbr:Che\_(2008\_film), z=dbr:Benicio\_del\_Toro)
% \begin{itemize}

    \item \textbf{Predicate grounding} (\textit{rdf:type}):
    LNN then proceeds to instantiate necessary facts at each input predicate that is part of the logic query, applying the global variable bindings obtained in the previous step. Often, especially for predicates pre-bound in the logic query, predicates may become fully bound at this step. While some fully bound SPARQL queries may seem redundant, a SPARQL query generally still needs execution to ascertain whether the predicate does satisfy the binding conditions, which is not necessarily always the case, such as when disjunctions are involved as in Figures~\ref{fig:lnn_shatner} and~\ref{fig:lnn_river}.
    
    For the predicate \textit{rdf:type}, the following SPARQL statement is queried: \texttt{SELECT ?x \{VALUES ?x dbr:Che\_(2008\_film)\} \{?x type dbo:Film\}}. This is an example of a fully bound query, and if \textit{rdf:type} satisfies the variable binding requirements, the fact (dbr:Che\_(2008\_film) rdf:type dbo:Film) is added for the predicate.
    
    \item \textbf{Predicate grounding} (\textit{dbo:country}):
    Similarly, for grounding the individual predicate \textit{dbo:country} a fully bound SPARQL query is executed, namely \texttt{SELECT ?x \{VALUES ?x dbr:Che\_(2008\_film)\} \{?x dbo:country dbr:Spain\}}. The KB confirms availability of the predicate, and the fact (dbr:Che\_(2008\_film) dbo:country dbr:Spain) is entered into \textit{dbo:country}. Again, although execution of this fully bound SPARQL query may seem redundant, there are situations where the truth of the embedded fact(s) needs confirmation, such as in the presence of disjunctions.
% \end{enumerate}
% \end{itemize}

\subsection{Reasoning methods}
Figure~\ref{fig:lnn_reasoning_methods} shows examples for LNN reasoning methods categorized under geographic and type-based reasoning.
Most KBQA datasets and the expressiveness of the associated ontologies can presently evaluate the following reasoning categories.

% \textbf{(1) Type-based reasoning:} Candidate queries can be eliminated based on inconsistencies with the type hierarchy in the knowledge base. For instance, ``Does Neymar play for Real Madrid" with faulty relation linking \textit{starring(Neymar, Real\_Madrid\_C)} will have incorrectly typed subject/object given predicate domain/range types.
% \begin{figure}[htpb!]
%   \centering
%     \includegraphics[width=0.5\textwidth,trim={0.05cm 0.6cm 18.9cm 0cm},clip]{images/lnn-cosmo.pdf}
%   \caption{The Cosmonauts example.} 
%   \label{fig:lnn_cosmo}
% \end{figure}
% \begin{figure}%[htpb!]
%   \centering
%     \includegraphics[width=0.5\textwidth,trim={0.05cm 0.2cm 18.9cm 0cm},clip]{images/lnn-shatner.pdf}
%   \caption{Multi-Hop reasoning LNN example.} 
%   \label{fig:lnn_shatner}
% \end{figure}
% \begin{figure}[htpb!]
%   \centering
%     \includegraphics[width=0.5\textwidth,trim={0.05cm 3.3cm 18.9cm 0cm},clip]{images/lnn-cuban.pdf}
%   \caption{The Cuban Missile Crisis example.} 
%   \label{fig:lnn_cuban}
% \end{figure}

% \textbf{(2) Geographic Reasoning:} As shown in Figure~\ref{fig:lnn_reasoning_methods}, identifying holonyms may correctly answer the question, potentially using the type-hierarchy.

% \textbf{(3) Temporal Reasoning:} A question that requires ordering of events based on Allen Algebra. For instance, ``Who is the youngest player" or ``Did X happen after Y?". \sysname'\texttt{s} LNN module performs type-based and geographic reasoning, while temporal reasoning is left for future work.

\subsubsection{Type-based reasoning}
\begin{figure}[t!]
  \centering
    \includegraphics[width=0.5\textwidth,trim={0cm 1.8cm 21.4cm 0cm},clip]{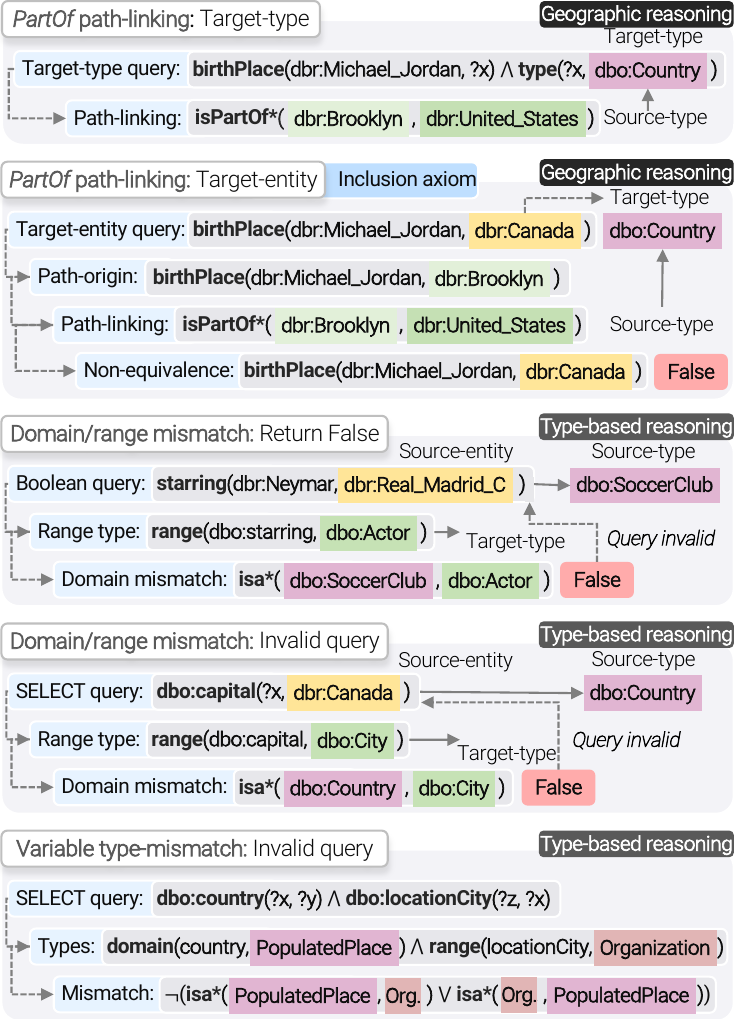}
  \caption{LNN reasoning methods.} 
  \label{fig:lnn_reasoning_methods}
\end{figure}

LNN performs type-based reasoning to discard queries inconsistent with the KB and likely to provide wrong answers, e.g. when relation linking gives \textit{dbo:starring} as predicate for ``Does Neymar play for Real Madrid", as shown in Figure~\ref{fig:lnn_reasoning_methods}. LNN expands queries to require correct range/domain typing for subject/object, e.g. \textit{dbo:starring(Neymar, Real\_Madrid\_C) $\land$ isa*(SoccerClub, Actor)} would resolve to \textit{False} even under the open-world assumption when hypernym transitivity fails to find \textit{SoccerClub} under \textit{Actor}. 

LNN eliminates type-invalid logic and then proceeds to evaluate alternative logic queries. Consider the example when the pipeline gives multiple options for ``Does Neymar play for Real Madrid": \textit{dbo:starring(Neymar, Real\_Madrid\_C)} and \textit{dbo:club(Neymar, Real\_Madrid\_C)}. 

Type-based reasoning would discard \textit{dbo:starring(Neymar, Real\_Madrid\_C)} and the next query option would be considered, namely \textit{dbo:club(Neymar, Real\_Madrid\_C)}.
Type-based reasoning is beneficially used to filter logic forms, although inconsistent KBs may result in discarding of otherwise viable logic queries.

\subsubsection{Geographic reasoning}
KBQA requires non-trivial reasoning such as type consistency and geographic reasoning. 
ASK questions that involve the retrieval of a fully specified triple undergo further LNN processing while the outcome is unknown under an open-world assumption. If no matching triple is found then a retrieval of available triples for the subject (or object) is performed for the specified predicate. The objective is then to ascertain whether any of the available objects (or subjects) are part of, or contained in the query-provided object. 

Predicates like \texttt{dbo:isPartOf} can help to determine inclusion, but if no match is found between available objects and the query-provided object, then an LNN support axiom is used. The matches provided by the inclusion predicates can be elevated to a holonym of similar type to the query-provided object, and an object-inclusion axiom helps to then infer that the matching object cannot be part of the query-provided object if it's already part of another object of the same type.

%---------------------------------

% \input{appendix}

\bibliographystyle{acl_natbib}
\bibliography{acl2021,ijcai21}